\documentclass[10pt,twocolumn,letterpaper]{article}

\usepackage{iccv}
\usepackage{times}
\usepackage{epsfig}
\usepackage{graphicx}
\usepackage{amsmath}
\usepackage{amssymb}
\usepackage{adjustbox}
\usepackage{subcaption}

\newcommand{\printfnsymbol}[1]{%
  \textsuperscript{\@fnsymbol{#1}}%
}

\usepackage[pagebackref=true,breaklinks=true,colorlinks,bookmarks=false]{hyperref}

\iccvfinalcopy 

\def\iccvPaperID{3957} 

\ificcvfinal\fi

\begin{document}

\title{Semantically Robust Unpaired Image Translation for Data with Unmatched Semantics Statistics}

\vspace{-0.2cm}
\author{
Zhiwei Jia\textsuperscript{1}\thanks{Co-first authors; work partly done during an internship at X. Code available at \href{https://github.com/SeanJia/SRUNIT}{https://github.com/SeanJia/SRUNIT}.} \quad 
Bodi Yuan\textsuperscript{2}\footnotemark[1] \quad 
Kangkang Wang\textsuperscript{2} \quad 
Hong Wu\textsuperscript{2} \quad \\
David Clifford\textsuperscript{2} \quad
Zhiqiang Yuan\textsuperscript{2} \quad 
Hao Su\textsuperscript{1} \\

\textsuperscript{1}UC San Diego {\tt\small \{zjia,haosu\}@eng.ucsd.edu} \\ 
\textsuperscript{2}X {\tt\small \{bodiyuan,kangkang,wuh,davidclifford,zyuan\}@google.com}
}
\vspace{-0.2cm}



\maketitle

\begin{abstract}
Many applications of unpaired image-to-image translation require the input contents to be preserved semantically during translations.
Unaware of the inherently unmatched semantics distributions between source and target domains, existing distribution matching methods (i.e., GAN-based) can give undesired solutions.
In particular, although producing visually reasonable outputs, the learned models usually flip the semantics of the inputs.
To tackle this without using extra supervisions, we propose to enforce the translated outputs to be semantically invariant w.r.t. small perceptual variations of the inputs, a property we call ``semantic robustness''.
By optimizing a robustness loss w.r.t. multi-scale feature space perturbations of the inputs, our method effectively reduces semantics flipping and produces translations that outperform existing methods both quantitatively and qualitatively.
\end{abstract}
\vspace{-0.2cm}

\section{Introduction} \label{sec:intro}
\vspace{-0.2cm}

\begin{figure}[t!]
  \centering
 \begin{minipage}{.48\textwidth}
   \centering
\includegraphics[width=0.8\linewidth]{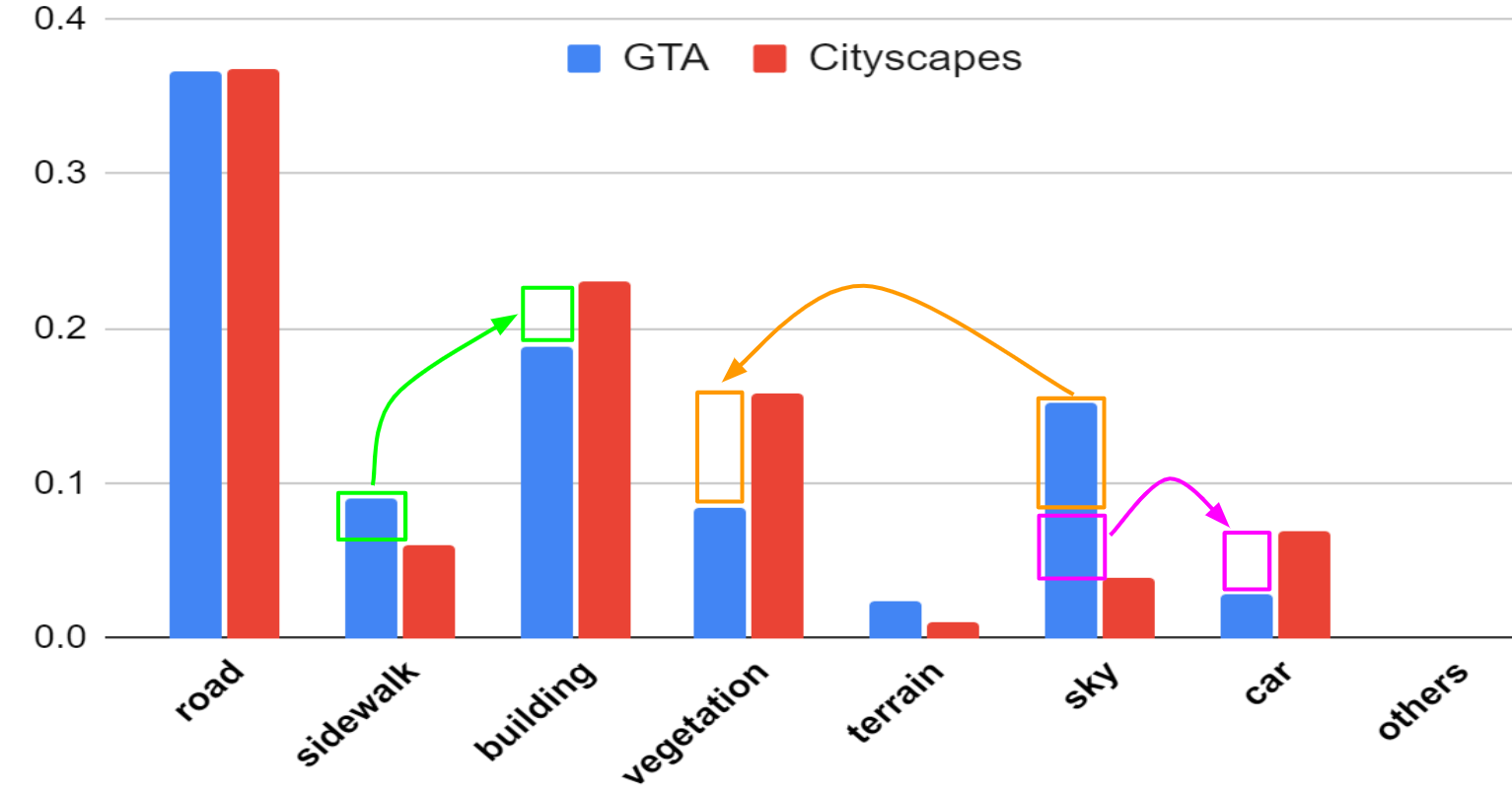}
    \vspace{-0.2cm}
  \caption[width=0.9\linewidth]{The class distributions in GTA vs. Cityscapes. During unpaired image translation, the generator has to flip the inputs' semantics to match the target distributions. Instances from over-represented semantic classes in the source domain (e.g., sky) can be flipped to those from underrepresented classes (e.g., vegetation).}
  \label{fig:1}
  \end{minipage}

 \begin{minipage}{.48\textwidth}
    \vspace{10pt}
    \centering
    \includegraphics[width=0.95\linewidth]{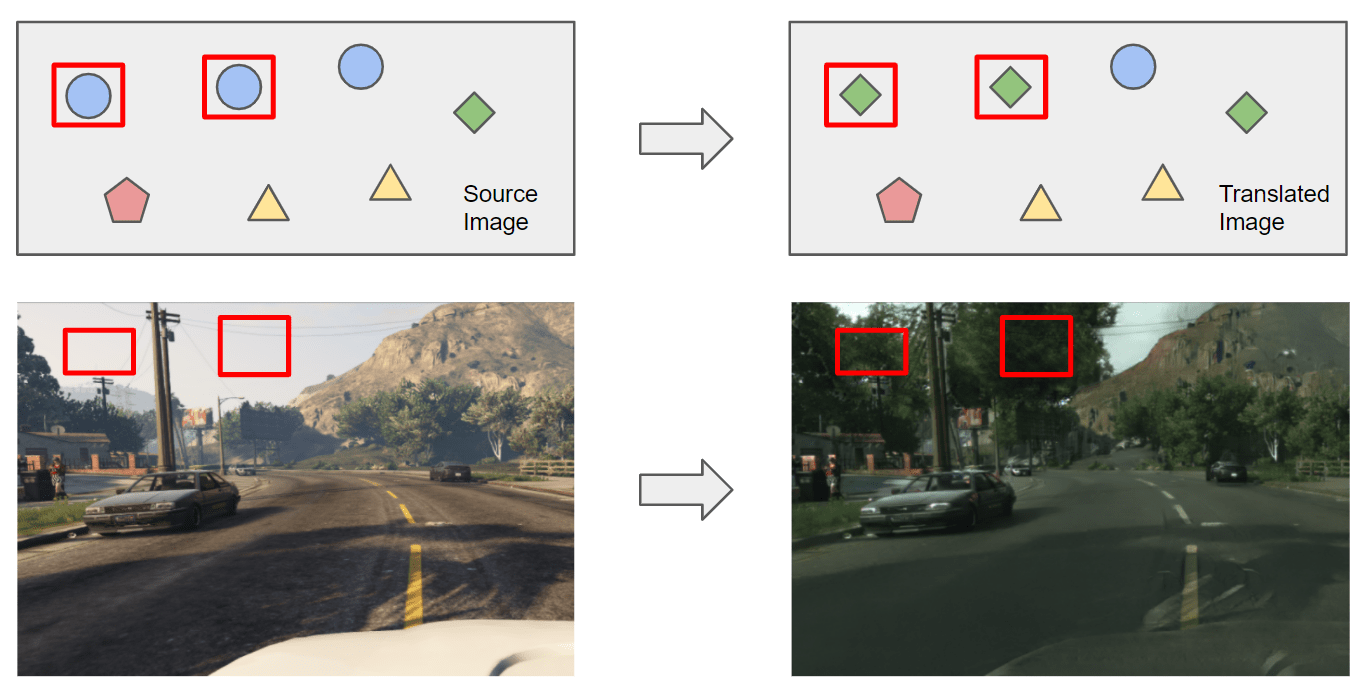}
      \caption{(\textbf{top}) Conceptually, forcing the distribution of translated images to match the target one causes  semantics (the different colored shapes) of the input images to get flipped.
      (\textbf{bottom}) An example of semantics flipping (highlighted in red boxes) from the GTA to Cityscapes task.} 
          \vspace{-0.3cm}

      \label{fig:2}
  \end{minipage}
\end{figure}
\vspace{10pt}

Recently, unpaired image-to-image translation \cite{chu2017cyclegan} has been very popular in the computer vision community.
Due to its general assumptions on the inputs (unlabelled images collected from different domains) and the easy accessibility of training data (does not use paired images), it is widely used in fields such as image manipulations, style transfer, domain adaptation, data augmentation, etc. \cite{donahue2018adversarial, zhang2017stackgan, reed2016generative, hoffman2018cycada, johnson2016perceptual, bousmalis2016domain, ledig2017photo, pathak2016context, samangouei2018defense, antoniou2017data}.
On the other hand, unpaired image translation remains a very challenging task owing to its unsupervised learning nature.
Without paired images that specify the exact domain mapping, one has to rely on visual cues to perform distribution matching (i.e., via GANs \cite{goodfellow2014generative}).
Existing GAN-based methods all rely on the adversarial loss that aims to optimally align image statistics between translation images and the target ones (in the marginal sense).
\emph{However, what if the two distributions should not be the same?}
In fact, the underlying distributions of semantics from the two domains are usually different, let alone the image distribution of translated images and target one.
We call this the unmatched semantics statistics problem, which is under-explored yet both critical and common for unpaired image translation tasks.

Similar to the language translation, the semantics of an image should be preserved during translations.
For instance, in the GTA to Cityscapes dataset \cite{cordts2016cityscapes}, while trees look different across domains, their identity/semantics remains the same.
In the Horse to Zebra task \cite{chu2017cyclegan}, a horse or zebra remains in the class Equus, not turning into a shack. 
Consider the translation as a two-stage process: firstly project an image from one domain to the shared semantics space, and then project it to the other domain.
When the source and target images are projected to the same semantic space and have different distributions in that space, we say the data have \emph{unmatched semantics statistics}.
Unpaired data from different domains generally have unmatched semantics statistics, unless they are very carefully constructed.
For example, in the Horse to Zebra dataset, there are more zebras than horses; in the GTA to Cityscapes dataset, more trees in Cityscapes than in GTA (see Fig. \ref{fig:1}).
Given unmatched semantics statistics, forcibly matching distributions between the translated and the target images can only give spurious solutions, where semantics get flipped only to match the target semantics statistics (see Fig. \ref{fig:2} for an example).
In Sec. \ref{sec:exp}, we demonstrate that semantics flipping is a critical and common issue in various GAN-based unpaired image translation frameworks. 

There are a few direct attempts at preserving the semantics during translations and thus reducing flipping. 
However, they either require extra supervision or pre-trained models \cite{hoffman2018cycada, taigman2016unsupervised} or are too restrictive (dataset-specific) and prone to artifacts \cite{benaim2017one, zhang2019harmonic, yang2020phase}.
In this paper, we propose to tackle this problem by encouraging that, during image translations, perceptually similar contents should be mapped to contents with high semantic similarity.
We call this property of the mapping as \emph{semantic robustness}.
In essence, semantic robustness ensures a consistent mapping that prevents the semantics of the inputs from being flipped easily. 
Specifically, based on the recently proposed framework CUT \cite{park2020contrastive},
we propose a semantic robustness loss w.r.t. multi-scale feature space perturbations of the input images.
We call our method SRUNIT (Semantically Robust Unpaired Image Translation) and empirically demonstrate its effectiveness in reducing semantics flipping.
SRUNIT outperforms existing GAN-based methods both qualitatively and quantitatively on several common datasets.


\section{Related Work} \label{sec:work}

\paragraph{Unpaired Image-to-image Translation}
Although lacking pixel-wise supervision, advances have been made in unpaired image-to-image translation by utilizing Generative Adversarial Networks (GANs) \cite{goodfellow2014generative}.
The central idea is to minimize the statistical difference (measured using the discriminators) between generated and target images by updating the generators.
These methods can roughly be sorted into two-sided methods such as \cite{li2018unsupervised, zhu2017unpaired, kim2017learning} and one-sided counterparts such as \cite{huang2018multimodal, liu2017unsupervised, liu2016coupled}.  

\vspace{-0.3cm}
\paragraph{Preserving Semantics in Image Translation}
More recently, efforts have been made in preserving the semantic content of the source images during the unpaired image translation.
There are several existing approaches.
Cycle consistency \cite{zhu2017unpaired} is proposed to enforce bijective mappings between domains so that semantic information will not be lost during translation.
Geometry consistency \cite{fu2019geometry} enforces equivariance of the generators regarding geometric transformations.
DistanceGAN \cite{benaim2017one} and HarmonicGAN \cite{zhang2019harmonic} encourage visual similarities within the source domain to be reflected in the target domain.
A spectral constraint based on the Fourier transform of the input images is proposed in \cite{yang2020phase}.
Attention-based methods \cite{mejjati2018unsupervised, tang2019attentiongan} are used to preserve the background during the translations.
Moreover, multiple work \cite{taigman2016unsupervised, bousmalis2017unsupervised, shrivastava2017learning, liang2017generative, park2020contrastive} adopt the idea that input and output images should be similar, measured by a function either pre-defined or learned contrastively.

\vspace{-0.3cm}
\paragraph{Robustness \& Generalization of DNNs}
Semantic robustness and semantics flipping discussed in our paper is related to both adversarial robustness and generalization ability.
Some work has tackled the adversarial robustness of GANs \cite{chu2017cyclegan, bashkirova2019adversarial, thekumparampil2018robustness}.
And some \cite{zhang2017discrimination, thanh2019improving, arora2017generalization} has explored their generalization properties.
In a broader context, both adversarial attack and defence have been extensively studied \cite{su2019one, gu2017badnets, huang2020one, szegedy2013intriguing, sun2019adversarial, prakash2018deflecting, goodfellow2014explaining, akhtar2018threat}, and many recent advances have been made for understanding the generalizability of DNNs \cite{neyshabur2015norm, sokolic2017robust, xu2012robustness, bartlett2017spectrally, zhou2018nonvacuous, dziugaite2017computing, arora2018stronger, jiang2018predicting, jia2019information}.

\section{Semantic Robustness} \label{sec:robust}
Many applications of unpaired image translation (style transfer, domain adaptation, data augmentation  \cite{hoffman2018cycada, johnson2016perceptual, antoniou2017data}) require the semantics of the inputs to be preserved during translations.
In this section, we will discuss the semantics flipping issue and the concept of semantic robustness.

\subsection{Unmatched Semantics Statistics} \label{sec:stats}


Most existing approaches for unpaired image-to-image translation do not explicitly study the mismatched distributions of the semantics across source and target domains.
This prevalent phenomena in unpaired translation tasks usually incur serious artifacts (see Fig. \ref{fig:2} bottom row).
To begin with, let us define the terms.
When translating images from one domain to the other, it is natural to assume an intermediate semantics space where resides the information to be preserved during translations. 
When converting images from a domain to the shared semantics space, we refer to the resulting distribution in this space as semantics distribution.
Due to the nature of unpaired image translation tasks where direct supervision of the paired relations is missing, we should assume that the unpaired data from different domains have different semantics distributions (i.e., the unmatched semantics statistics).
Most widely available datasets fall into this category (e.g., see Fig. \ref{fig:1} for unmatched semantics statistics between GTA \cite{richter2016playing} and Cityscapes \cite{cordts2016cityscapes}).
A few exceptions are those originally constructed for paired translation (e.g., Maps to Photos \cite{isola2017image}).

\subsection{The Semantics Flipping Issue} \label{sec:issue}
We argue that provided the unmatched semantics statistics between source and target domains, an inherent problem for GAN-based unpaired image translation frameworks is the semantics flipping issue.

The central idea of GAN-based methods is to match the image statistics between translated images and target images as much as possible.
Multiple metrics for evaluating the translation performance follow this principle, i.e. they measure some sort of statistical distance between generated and target images (FID, MMD, etc. \cite{heusel2017gans, gretton2012kernel}).
This is indeed problematic, as the generated and the target distribution should not be the same, provided that the source and target domains have discrepancies in semantics statistics.
We observe that the learned translation models by existing methods usually are undesired solutions (e.g., Fig. \ref{fig:2} bottom row), which, although producing visually reasonable outputs, systematically flip contents into other semantics.
This is because only through semantics flipping can the generators produce images that match the statistics of the target domain (see Fig. \ref{fig:1} as an illustration of this process).

\subsection{Limitations of Existing Approaches} \label{sec:limit}
Most existing unpaired image translation frameworks do not explicitly tackle and, in fact, suffer from the semantics flipping issue (empirically demonstrated in Sec. \ref{sec:exp}).
For two-sided domain mapping methods, cycle consistency \cite{zhu2017unpaired} are the most popular technique which suggests using bijective (and therefore information-preserving) mappings.
However, as pointed out in \cite{chu2017cyclegan}, CycleGAN can learn to hide information in plain sight such that semantics flipping still occurs while the information is preserved during the translations.
One-sided domain mapping approaches directly pose constraints on the generators to preserve meaningful information.
GcGAN \cite{fu2019geometry} proposes geometry consistency to enforce that the translation functions are equivariant w.r.t. common geometric transformations.
However, spurious solutions with semantics flipping can also be equivariant as such.
Another line of work is to enforce some sort of relations between input and output images (or image patches), for example, by perceptual similarity or statistical dependency \cite{liang2017generative, park2020contrastive}.
Since these methods have their correspondence learned unsupervisedly (e.g., contrastively), its inaccuracy can lead to spurious enforcement with more semantics flipping (or artifacts).
Alternatively, \cite{yang2020phase} uses a spectral constraint to maintain semantics.
The approach might fail in general and was only shown to be successful in translation tasks across visually similar domains.
Although methods with ground truth perceptual similarity can reduce semantics flipping \cite{hoffman2018cycada, taigman2016unsupervised}, they require extra supervision or pre-trained models that are not available for a general unpaired image translation task.

\vspace{-0.1cm}
\subsection{Semantic Robustness to the Rescue}
\label{sec:rescue}
Other than directly enforcing relations between input and output images, we propose to encourage that small perceptual variation of the input images (or patches) should not change the semantics of the corresponding transformed images (or patches).
We call this property of the generators as \emph{semantic robustness}.
Notice that the perceptual similarity between images (or patches) refers to the distance measured in the feature space (e.g., CNN features of the images), rather than in the raw pixel space.
We argue that increasing the semantic robustness of the generators can effectively reduce semantics flipping during translations.
Intuitively, an input image (or patch) should have its semantics invariant under small perceptual perturbations, and thus, the semantics of the corresponding translated image (or patch) should also be invariant.
Remember that semantics flipping happens as the generator is forced to match the target statistics by transforming semantically over-represented contents from the source domain to less represented ones (see Fig. \ref{fig:1}).
\emph{Semantic robustness encourages a consistent translation such that contents of the same semantics are not transformed into contents of several different semantics.}
As a result, it prevents forceful distribution matching and mitigates the flipping issue.

How do we obtain the semantics from images in the first place?
Without relying on extra supervision or pre-trained models, contrastive learning approaches (e.g., \cite{park2020contrastive}) can learn to extract features that are domain-invariant, which we consider as the semantics of the inputs.
One might find it intuitive to directly enforce that the translation should not change these ``semantics'' of the input to reduce semantics flipping.
However, this direct approach does not work well (see our ablation study in Sec. \ref{sec:ablation}).
Interestingly, these extracted semantics can be used to effectively reduce flipping by instead enforcing semantic robustness (i.e., the semantics of the translated image be invariant to perceptual variations of the inputs).
This is partly because the latter indirect constraint is a ``soft'' version of the former direct constraint and is more robust w.r.t. the inaccuracy of the extracted semantics which is contrastively learned.

\section{Method} \label{sec:srunit}

\subsection{Preliminary: CUT} \label{sec:cut}
The goal of unpaired image-to-image translation is to learn functions between two domains $X$ and $Y$ given training samples $\left\{x_{i}\right\}_{i=1}^{N}$, $\left\{y_{j}\right\}_{j=1}^{M}$ sampled from $p_X(x)$ and $p_Y(y)$.
Recently, several one-sided methods were proposed, 
which essentially learn the generator $G: X \rightarrow Y$ and the discriminator $D_{Y}$ that aims to distinguish between images $\{x\}$ and translated images $\{F(y)\}$.
Commonly, the training objective consists of multiple pieces.
The first one is the adversarial losses \cite{goodfellow2014generative}, Eqn. \ref{eqn:1}, for matching the marginal distribution of generated images to that of the target images.
\begin{align} \label{eqn:1}
\mathcal{L}_{\mathrm{GAN}}(G, &D_{Y}, X, Y) =\mathbb{E}_{y \sim p_{Y}(y)}\left[\log D_{Y}(y)\right] \nonumber \\
&+\mathbb{E}_{x \sim p_{X}(x)}\left[\log \left(1-D_{Y}(G(x))\right]\right. 
\end{align}
The second part is usually a loss constraining the generator $G$ to perverse desired contents during translations.
For instance, the recent state-of-the-art method Contrastive Unpaired Translation (CUT) \cite{park2020contrastive} tries to maximize the mutual information between the input and generated output via contrastive learning.
It utilizes InfoNCE loss \cite{oord2018representation} to learn an embedding that associates corresponding patches (of input and translated images) to each other while disassociating them if otherwise.
By doing so, it learns encoders that extract domain-invariant features of the input images at multiple scales.
At each scale, the feature (an $\mathbb{R}^{256}$ vector) at one position from the input image is denoted as ``query'' $v$; the corresponding feature in the translated image is denoted the ``positives'' $v^+$; the features at $N$ other locations from the input images are the ``negatives'' $v^{-}$.
Formally, the contrastive loss is set up as an $(N + 1)$–way classification as below (where $\tau$ is the temperature).
\begin{align} \label{eqn:2}
&\ell\left(\boldsymbol{v}, \boldsymbol{v}^{+}, \boldsymbol{v}^{-}\right) =  \\
&-\log \left[\frac{\exp \left(\boldsymbol{v} \cdot \boldsymbol{v}^{+} / \tau\right)}{\exp \left(\boldsymbol{v} \cdot \boldsymbol{v}^{+} / \tau\right)+\sum_{n=1}^{N} \exp \left(\boldsymbol{v} \cdot \boldsymbol{v}_{n}^{-} / \tau\right)}\right]\nonumber
\end{align}


Although encouraging semantic correspondence between input and output images, CUT still suffers from semantics flipping when the two domains have unmatched semantics statistics.
This is because the contrastively learned semantics are not accurate enough to ensure successful correspondence enforcement across domains.
Nevertheless, when combined with other techniques to improve semantic robustness, these semantics extractors can be used to successfully reduce semantics flipping. 

\begin{figure*}[t!]
  \centering
    \includegraphics[width=\linewidth]{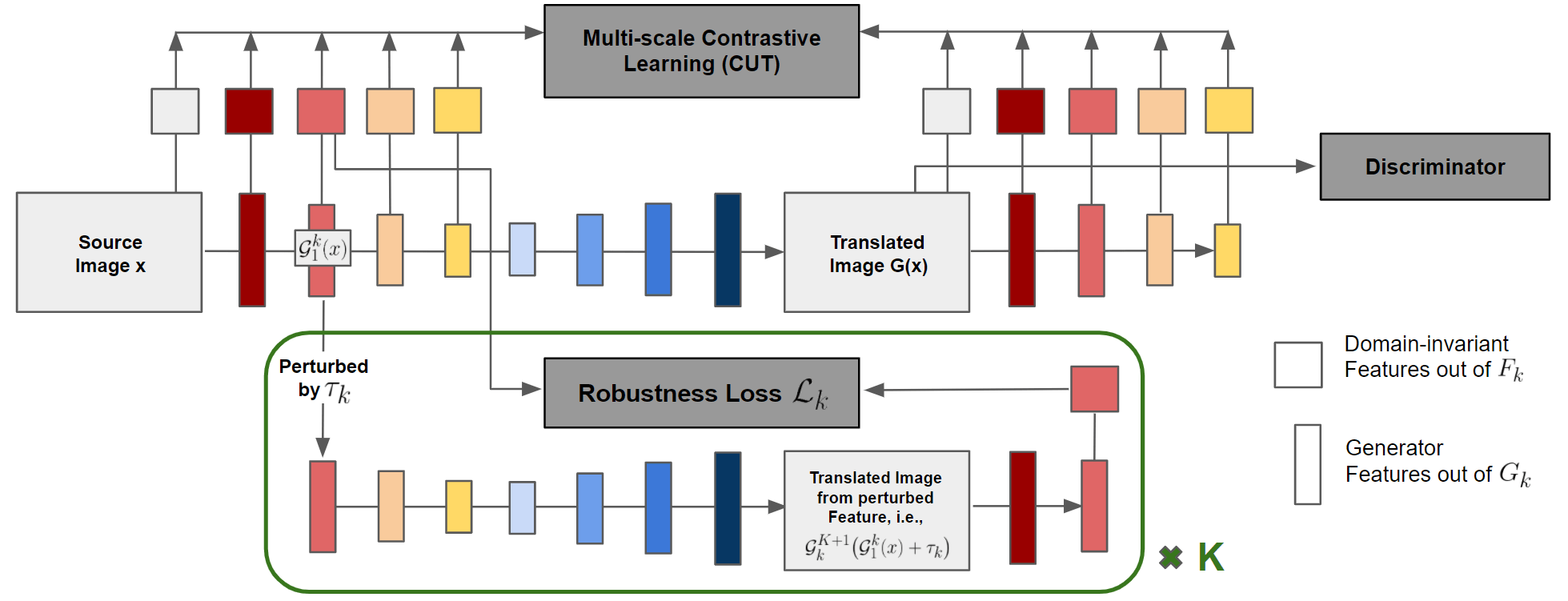}
  \caption{Our method improves semantic robustness by making the semantics of translated output invariant to small feature space variations of the inputs. The contents in the green box exemplify the data-flow for $\mathcal{L}_k$ at one specific scale $k$. There are in total $K$ such losses; each corresponds to one of the $K$ selected layers in the generator, respectively.}
  \label{fig:4}
  \vspace{-0.3cm}
\end{figure*}

\subsection{Semantically Robust Unpaired Image Translation (SRUNIT)} \label{sec:method}

Our method is based on CUT \cite{park2020contrastive} and the semantic robustness we proposed in Sec. \ref{sec:rescue}.
As illustrated in Fig. \ref{fig:4}, in CUT, $K$ layers (denoted $\{G_k\}$), including the input layer $G_1$ (an identity function), are selected from the first half of the generator $G$.
Together with the rest of the network, denoted $G_{K+1}$, we have $G = G_{K+1} \circ G_K \circ \dots \circ G_1$.
We further define $\mathcal{G}_i^j = G_{j}\circ\dots\circ G_{i+1}$ as a forward propagation via $(j-i)$ components of $G$.
For instance, $G(x) = \mathcal{G}_1^{K+1}(x)$.
During CUT training, at each scale $k \in \{1...K\}$ (just like in CUT, there are $K$ scales in total), a feature extractor $F_k$ that consumes the output of $G_k$ (a layer in the generator) is learned by Eqn. \ref{eqn:2}, where $v, v^{-}$ and $v^{+}$ are the outputs of $F_k$.
The optimization of Eqn. \ref{eqn:2} encourages the feature $F_k(\mathcal{G}_1^k(G(x)))$ to remain close to $F_k(\mathcal{G}_1^k(x))$.

We consider the perceptual variations mentioned in our semantic robustness concept as the random perturbations in the output space of $G_k$, and consider outputs of $F_k$ as the semantics (by Eqn. \ref{eqn:2} $F_k$ extracts domain-invariant features).
Then, we propose to improve semantic robustness by minimizing the loss $\mathcal{L}_{robust} = \frac{1}{K}\sum_{k=1}^K \mathcal{L}_k$, where
\begin{align} \label{eqn:3}
    \mathcal{L}_k = \mathbb{E}_x \Big[\frac{1}{||\tau_k||_2} \Big\Vert &F_k(\mathcal{G}_1^k(x)) \ - \\  &F_k(\mathcal{G}_1^k(\mathcal{G}_k^{K+1}(\mathcal{G}_1^k(x)+\tau_k))) \Big\Vert_2 \Big] \nonumber
\end{align}
The $\tau_k$ refers to some random perturbation.
As shown in Fig. \ref{fig:4}, $\mathcal{L}_k$ measures the distance between extracted semantics at scale $k$ from the input image and that of the corresponding translated image under feature space perturbation to the input.
We can see that minimizing $\mathcal{L}_k$ indirectly enforces semantic robustness, which is the condition that ``transformed images should have their semantics invariant to small feature space variations of the inputs''.
Formally, this condition can be measured by
\begin{align} \label{eqn:4}
    \mathbb{E}_x \Big[\frac{1}{||\tau_k||_2} \Big\Vert &F_k(\mathcal{G}_1^k(G(x))) \ - \\  &F_k(\mathcal{G}_1^k(\mathcal{G}_k^{K+1}(\mathcal{G}_1^k(x)+\tau_k))) \Big\Vert_2 \Big] \nonumber
\end{align}
$\mathcal{L}_k$ and Eqn. \ref{eqn:4} are closely related by the triangle inequality since, by contrastive learning (Eqn. \ref{eqn:2}), we have $F_k(\mathcal{G}_1^k(G(x)))$ remain close to $F_k(\mathcal{G}_1^k(x))$.
In fact, our approach produces better translation results than directly optimizing Eqn. \ref{eqn:4}, because the latter can harm the diversity of the translation (the mode collapse issue).
Optimizing $\mathcal{L}_k$ can be seen as an adaptive version of optimizing Eqn. \ref{eqn:4}, adjusted by the distance between $F_k(\mathcal{G}_1^k(G(x)))$ and $F_k(\mathcal{G}_1^k(x))$.
See Sec. \ref{sec:ablation} for empirical evidence and see Appendix for a detailed discussion.

One might ask a question: why not directly minimize the distance between $F_k(\mathcal{G}_1^k(G(x)))$ and $F_k(\mathcal{G}_1^k(x))$ to enforce semantics-preserving translations.
This will make duplicate efforts, similar to the contrastive loss (Eqn. \ref{eqn:2}) used in CUT.
We show in the ablation study (Sec. \ref{sec:ablation}) that doing so can actually hurt the performance.

Moreover, we adopt the patch-based approach so that $x$ refers to the input image patches.
In practice, one can choose to only include a random subset of $\{\mathcal{L}_k\}$ in each training iteration to reduce the computational complexity of optimizing $\mathcal{L}_{robust}$.
See Sec. \ref{sec:impl} for more details.

\subsection{Advantages Over Distance-Preserving Methods} \label{sec:adv}
Our semantic robustness approach is advantageous to the distance-preserving approach (e.g., DistanceGAN \& HarmonicGAN \cite{benaim2017one, zhang2019harmonic}), which aims to improve the semantic consistency of the translations by maintaining
the distance between different parts of the same sample during the transformations.
See Sec. \ref{sec:ablation} for the empirical results.

Firstly, assuming no access to extra supervision or pre-trained models, the distance used in this line of work is based on image pixels, for instance, the (standardized) L1 distance between raw pixels or color histograms.
Compared to the CNN feature-based perceptual similarity used in the concept of semantic robustness, pixel space similarity metrics are far more sensitive to geometric transformations, change of lighting conditions, etc.; thus, they usually fail to capture the essential information in the images.
The authors of HarmonicGAN provide an option to use CNN features from pre-trained models to measure visual distance, which is domain-specific and requires prior knowledge, making this approach restrictive.
Ours utilizes the contrastively learned features, being both universal and effective.

Secondly, the underlying principle of the distance-preserving approach is usually violated, i.e., the visual distance between contents in the source domain become changed when translated to the target domain.
For instance, in the Label to Image task from Cityscapes \cite{cordts2016cityscapes} (i.e., translating semantic labels as inputs to street-view images), 
two identical image patches of the same semantics should not be mapped to visually identical outputs.  
Forcefully maintaining the visual distance as such leads to serious artifacts and hurts the diversity of the translations.
Instead, our method encourages the outputs to be semantically the same, which can still be diverse and high-quality.

Thirdly, our method focuses on improving the robustness of the translations from the source domain w.r.t. feature space perturbations in all directions.
Whereas, distance-preserving methods do so only w.r.t. pixel space perturbations in the direction of other source images (or in practice, in the direction of other patches from the same source image).
This makes our approach much more efficient and effective in reducing semantics flipping.

\vspace{-0.1cm}
\section{Experiments} \label{sec:exp}
In this section, we demonstrate how our method (denoted SRUNIT) effectively reduces semantics flipping and produces translations that outperform existing approaches both quantitatively and qualitatively for several popular unpaired image translation tasks.
In specific, we compare SRUNIT with CycleGAN \cite{zhu2017unpaired}, GcGAN \cite{fu2019geometry}, DRIT \cite{lee2018diverse} and CUT \cite{park2020contrastive}.
Notice that some popular datasets are designed for paired translation tasks and are not very realistic for unpaired image translation (they admit perfectly aligned semantics statistics across the domains).
Therefore, we sub-sample them (which aggravates the flipping issue) so that the setup becomes more realistic.  

\subsection{Quantitative Evaluation} \label{sec:quan}

It is critical to choose the right metrics to quantitatively evaluate the translation performance, as the focus of this paper is on reducing semantics flipping.
Popular metrics such as FID and MMD \cite{heusel2017gans, gretton2012kernel} ignore the unmatched semantics statistics nature of unpaired image translation datasets (see the discussion in Sec. \ref{sec:issue}); consequently, they are not suitable and can be even misleading here.
Instead, we use datasets where (partial) information of the ground truth translated results are available and use corresponding metrics for evaluation. 
Some datasets (e.g., Aerial Photo to Google Map) directly provide ground truth correspondence, easing the evaluation of the translation quality.
Others (Label to Image, GTA to Cityscapes, etc.) do not have such ``ground truth'' translation.
On these datasets, we follow the common practice to compute metrics based on pre-trained models \cite{chu2017cyclegan,fu2019geometry,park2020contrastive}.
The intuition is that the more accurate the models (trained on source images) can classify the target images, the better these generated images are \cite{isola2017image}. 
\vspace{-0.3cm}

\begin{table*}[h]
\begin{adjustbox}{width=2\columnwidth,center}
\begin{tabular}{ c|ccc|ccc|ccc|ccc} 
 \hline
  & \multicolumn{3}{c|}{Label → Image} & \multicolumn{3}{c|}{GTA → Cityscapes} & \multicolumn{3}{c|}{Map → Photo} & \multicolumn{3}{c}{Photo → Map} \\

 \hline
 method & pxAcc & clsAcc & mIoU & pxAcc & clsAcc & mIoU & Dist & $\textrm{Acc}(\delta_1)$ & $\textrm{Acc}(\delta_2)$ & Dist & $\textrm{Acc}(\delta_1)$ & $\textrm{Acc}(\delta_2)$  \\
 \hline
 CycleGAN & 66.36 & 27.24 & 21.31 & 66.33 & 32.53 & \textbf{23.84}   & 70.16 & 28.67 & 43.88 & 23.02 & 16.11 & 32.65  \\
 GcGAN & 65.30 & 27.78 & 21.41 & 65.62 & 32.38 & 22.64 & 71.47 & 28.87 & 43.48 & 23.62 & 15.00 & 30.65  \\ 
 DRIT & 72.74 & 28.13 & 22.06 &  64.28 & 32.17 & 20.99      & 70.87 & 28.97 & 43.56 & 24.19 & 13.94 & 29.01  \\
 CUT & 75.09 & 29.70 & 23.43 & 64.59 & 32.19 & 20.35 & 70.28 & 28.86 & 44.07 & 23.44 & 16.25 & 31.34  \\ 
 SRUNIT (ours) & \textbf{80.70} & \textbf{33.95} & \textbf{27.23} & \textbf{67.21} & \textbf{32.97} & 22.69 & \textbf{68.55} & \textbf{30.41} & \textbf{45.91} & \textbf{23.00} & \textbf{17.67} & \textbf{32.78}  \\ 
 \hline
\end{tabular}
\end{adjustbox}
\caption{Average pixel prediction accuracy (pxAcc), average class prediction accuracy (clsAcc), Mean IoU (mIoU), Average L2 distance (Dist) and pixel accuracy with threshold (Acc) measured for the tasks. The best entries are highlighted in bold.}
\label{tab:1}
\end{table*}

\begin{figure}[t!] 
	\centering
	\includegraphics[width=0.47\textwidth]{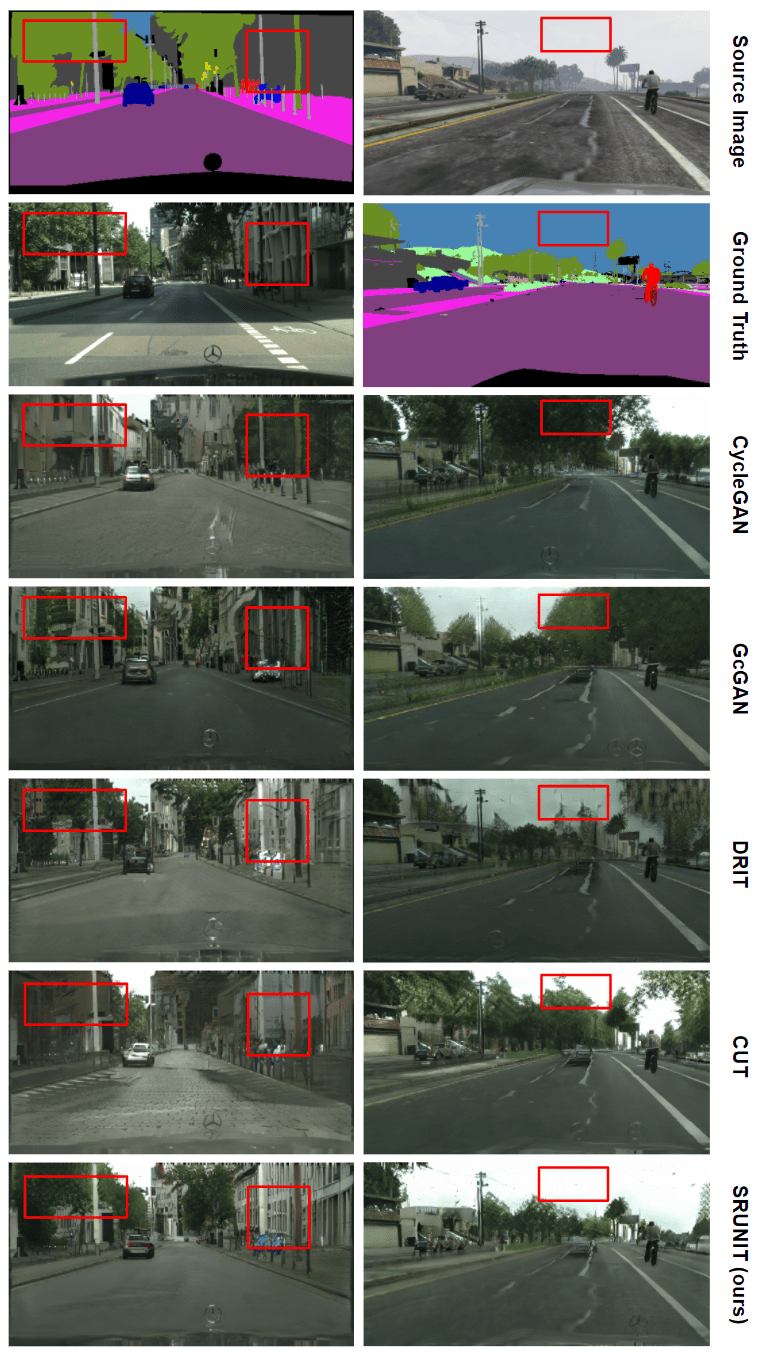}
    \vspace{-0.2cm}
	\caption{Visual results of the Label to Image and GTA to Cityscapes tasks (first and second column, respectively). Row 2 of column 2 shows the ground truth mask as no such ground truth image exists. Although not solved perfectly, semantics flipping is effectively reduced by our method (some improvements are highlighted in red boxes).}
    \label{fig:5}
    \vspace{-0.7cm}
\end{figure}

\begin{figure*}[t!]
	\centering
	\includegraphics[width=0.95\textwidth]{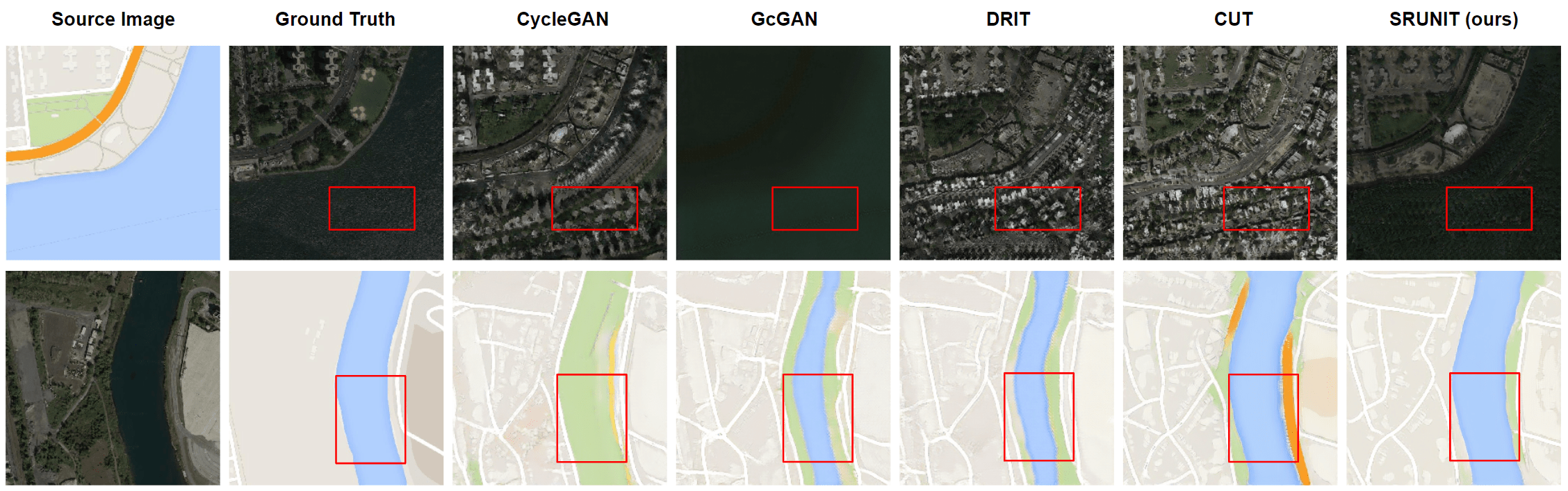}
	\caption{Visually in the Photo to Map tasks, our method effectively reduces semantics flipping (highlighted in red boxes).}
    \label{fig:6}
     \vspace{-0. cm}
\end{figure*}

\vspace{-0.2cm}
\subsubsection{Cityscapes Label → Image}

Cityscapes \cite{cordts2016cityscapes} is a real-world image dataset popular for benchmarking semantic segmentation and image translation.
The dataset is originally constructed for paired translation.
To ensure a reasonable level of unmatched semantics statistics between the two domains, we sub-sample around 1500 images from the RGB semantic label images and 1500 from the street-view images according to K-means clustering results based on histograms of the semantic labels.
Each image is resized to $512 \times 256$ and during training, we randomly crop the $256 \times 256$ patches.
As a result, the two domains have unmatched semantics statistics.
We use the 500 validation set in Cityscapes for evaluation (whose ground truth semantic labels are provided).
We use three metrics (as in \cite{park2020contrastive}) to provide a comprehensive evaluation of the translation quality.
They are the mean pixel accuracy, class accuracy (i.e., class-weighted pixel accuracy) and mean IoU (default metric for semantic segmentation).
These metrics are computed by using a light-weight publicly available DeepLab V3 \cite{chen2017rethinking} model pre-trained on the Cityscapes semantic segmentation task (refer to the Appendix for details).
We notice that, for these datasets, there are no standard pre-trained models for evaluation across existing work (e.g., CUT uses DRN \cite{yu2017dilated} and CycleGAN \& GcGAN use FCN \cite{long2015fully}).
We choose DeepLab V3 because its pre-trained models are public available and it is in general a better model for semantic segmentation.
Table \ref{tab:1} and Fig. \ref{fig:5} show that SRUNIT produces results better than existing methods by a large margin.
All detailed information is in the Appendix.

\subsubsection{GTA → Cityscapes}
GTA5 \cite{richter2016playing} is another popular dataset of 24966 synthesized images from the game Grand Theft Auto 5.
We set aside 500 images from GTA5 for evaluation and use the remaining ones together with all the 2975 Cityscapes images (from Cityscapes' fine-labeled training set) for training.
Similar to the Label to Image task, we resize all images to $512 \times 256$ and randomly crop $256 \times 256$ patches for training.
The two datasets have quite different semantics statistics (as displayed in Fig. \ref{fig:1}).
Again, we use the DeepLab model to compute the three metrics.
The quantitative results in Table \ref{tab:1} and qualitative results in Fig. \ref{fig:5} demonstrate the effectiveness of our proposed SRUNIT.

\subsubsection{Google Map → Aerial Photo}
The Google Maps dataset \cite{isola2017image} contains in total 2194 (map, aerial photo) pairs of images around New York City and is widely used in both paired and unpaired image translation \cite{isola2017image, zhu2017unpaired, fu2019geometry}.
The dataset is split into 1096 pairs and 1098 pairs for training and test sets, respectively.
Since it is original constructed for paired image translation, we sub-sample map images and aerial photos from the training set (around 600 from each domain) so that there is a reasonable amount of difference between the semantics statistics from the two sets.
We do so by K-means clustering the color histogram of the images (see the Appendix for details).
We use all the 1098 test set pairs for evaluation.
Images are resized to $256 \times 256$ to accommodate all methods in our comparison.
We measure the quality of translation by average pixel L2 distance (Dist) and pixel accuracy (\%), following \cite{fu2019geometry}, where given a ground truth pixel $p_i = (r_i,g_i,b_i)$ and the prediction $p_i' = (r_i',g_i',b_i')$, the accuracy of $p_i'$ is computed as 1 if $\max \left(\left|r_{i}-r_{i}^{\prime}\right|,\left|g_{i}-g_{i}^{\prime}\right|,\left|b_{i}-b_{i}^{\prime}\right|\right)<\delta$ and 0 otherwise:
We use $\delta = 30, 50$ as the domain of aerial photos has large diversity. 
Table \ref{tab:1} and Fig. \ref{fig:6} demonstrate the clear advantages of our method over existing ones.

\vspace{-0.2cm}
\subsubsection{Aerial Photo → Google Map}
The evaluation protocol is similar as above, except that we use smaller $\delta = 3, 5$ since the domain of google maps has much less diversity than the other domain.
Again, Table \ref{tab:1} and Fig. \ref{fig:6} show the advantage of our approach.

\begin{figure*}[t!]
	\centering
	\includegraphics[width=0.95\textwidth]{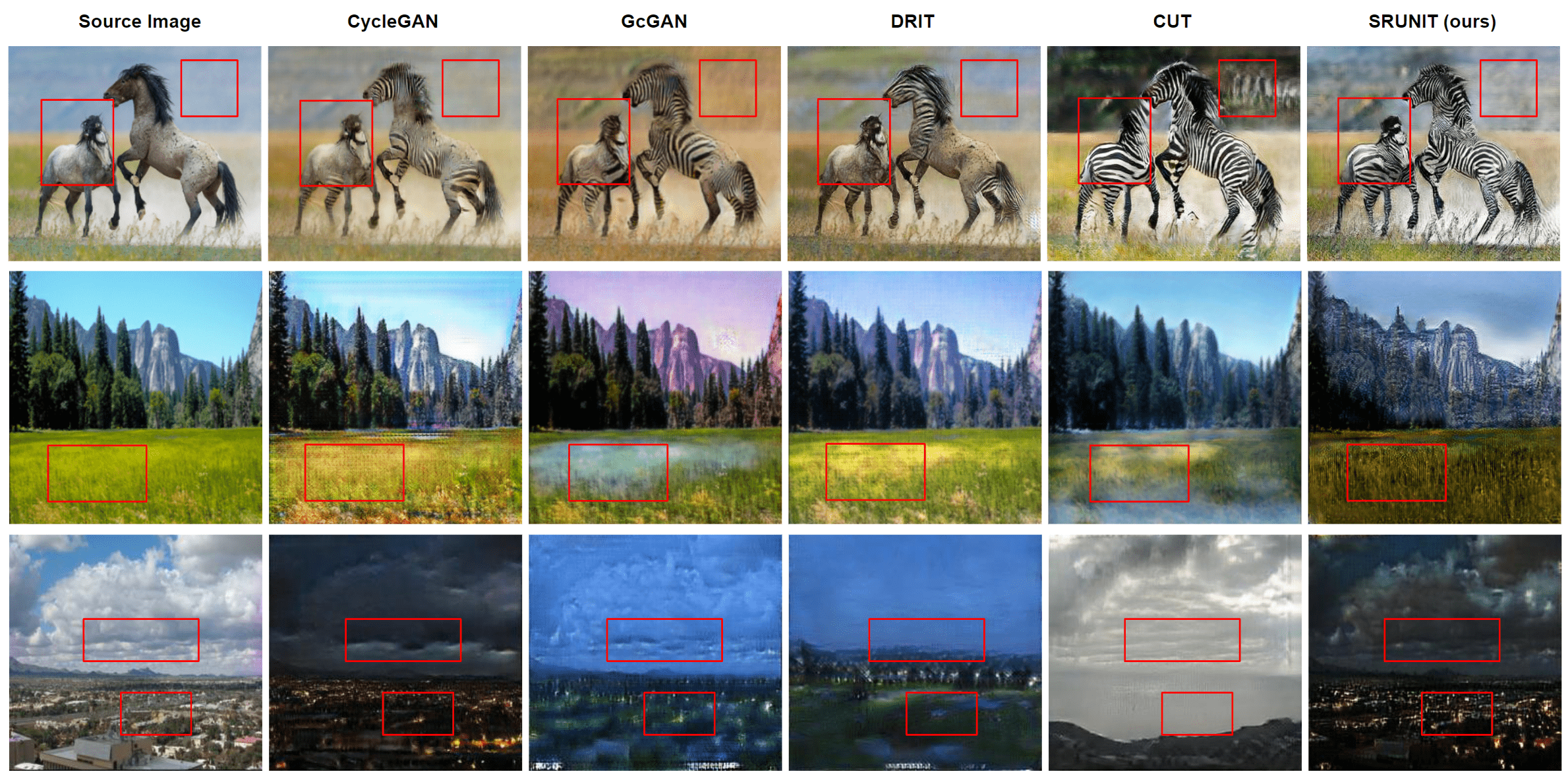}
	\caption{Visual comparisons on the three datasets (Horse to Zebra, Summer to Winter, Day to Night). Semantics flipping (or, in general, areas where our method improves over others) are highlighted in red boxes. See Appendix for more samples.}
    \label{fig:7}
    \vspace{-0.5 cm}
\end{figure*}

\subsection{Qualitative Evaluation} \label{sec:qual}
Besides the aforementioned tasks, we show more visual results on the following three popular datasets (all training images resized to $256 \times 256$).
In Fig, \ref{fig:7}, we demonstrate that SRUNIT produce images of better or comparable quality compared to others.
Due to the lack of ground truth translation information, it is not possible to quantitatively measure how well our model reduces semantics flipping.

\vspace{-0.2cm}

\paragraph{Horse → Zebra} A famous dataset consists of 1067 and 1334 training images for horses and zebras, respectively  (\href{https://www.tensorflow.org/datasets/catalog/cycle_gan#cycle_ganhorse2zebra}{link}).
The two domains have different semantics statistics.

\vspace{-0.5cm}

\paragraph{Summer → Winter} A dataset of photos of Yosemite constructed by authors of CycleGAN. The training set consists of 1231 summer images and 962 winter images (\href{https://www.tensorflow.org/datasets/catalog/cycle_gan#cycle_gansummer2winter_yosemite}{link}).
Again, the two domains have different semantics statistics.

\vspace{-0.5cm}

\paragraph{Day → Night} A dataset of outdoor scenes used in \cite{isola2017image, fu2019geometry}.
While the original dataset consists of paired images,
we sub-sample 1418 day images and 391 night images so that the semantics statistics are different (details in Appendix).

\subsection{Implementation details} \label{sec:impl}


We follow CUT \cite{park2020contrastive} for the choice of network architecture and the training setup (learning rate, number of epochs, etc.).
The $\tau_k$ used in Eqn. \ref{eqn:3} is sampled independently as a vector for each coordinate of the feature maps produced by $G_k$.
We first project standard multivariate Gaussian random variable into the unit sphere and then resize it with its magnitude sampled uniformly in $[10^{-7}, T]$, where the default $T$ we choose is $0.1$ (we fine-tune it in $[0.01, 0.2]$).
We set the default coefficient of the loss term $\mathcal{L}_{robust}$ as $10^{-4}$, and fine-tune it in $[10^{-5}, 10^{-3}]$.
By default we compute $\{\mathcal{L}_k\}$ (see Eqn. \ref{eqn:3}) for all $5$ feature extractors $\{F_k\}$ used in the CUT paper.
We fine-tune it by leaving one $\mathcal{L}_k$ out for each variant. 
We find all these hyper-parameters relatively robust.
Optimizing $\mathcal{L}_{robust}$ might lead to training instability at the beginning of the adversarial training; thus we only apply it after finishing $\frac{1}{4}$ of the total epochs.
See Appendix for full implementation details.




\section{Ablation Study} \label{sec:ablation}
Here we justify the design choice of our proposed semantic robustness loss $\mathcal{L}_{robust}$.
We perform all the following experiments on the Label to Image dataset as it is a relatively challenging task. 
We use CUT as the backbone (as used in SRUNIT).
The results are shown in Table \ref{tab:2}.

Firstly, we show our method's advantage over the distance preserving approach (see also Sec. \ref{sec:adv} for a discussion).
We train a model denoted E1 by adding the self-distance constraints from DistanceGAN \cite{benaim2017one} to the CUT backbone and E2 by adding a patch-based distance-preserving constraint in the style of HarmonicGAN \cite{zhang2019harmonic} (not exactly the same, though).
To verify the necessity of using feature extractor $F_k$ in $\mathcal{L}_{robust}$, we train E3 by removing the function calls to $F_k$ in Eqn. \ref{eqn:3}.
To show that Eqn. \ref{eqn:3} is a better proxy of Eqn. \ref{eqn:4} (as discussed in Sec. \ref{sec:method}), we train E4 by using Eqn. \ref{eqn:4} instead of Eqn. \ref{eqn:3} when optimizing the $\mathcal{L}_{robust}$.
We train a model E5 to illustrate that directly minimizing the distance between semantics (extracted by $F_k$) of the input and that of the output does not work (also see Sec. \ref{sec:method} for a discussion).
We further show that applying constraints on the discriminator instead of the generator is not a better way to improve the semantic robustness of the model.
We do so by training a model E6 with Lipschitz penalty \cite{gulrajani2017improved} on the discriminator in the spirit of WGAN \cite{gulrajani2017improved}.
We provide full details in the Appendix.

\begin{table}[h]
\begin{adjustbox}{width=\columnwidth,center}
\begin{tabular}{c|ccccccc} 
 \hline
  & E1 & E2 & E3 & E4 & E5 & E6 & SRUNIT \\
 \hline
 pixAcc & 74.46 & 75.31 & 75.42 & 76.38 & 74.86 & 76.25 & \textbf{80.70}  \\
 clsAcc & 29.84 & 30.13 & 30.43 & 31.13 & 29.79 & 30.92 & \textbf{33.95}  \\ 
 mIoU & 23.52 & 23.86 & 23.89 & 24.71 & 23.22 & 23.51 & \textbf{27.23} \\
 \hline
\end{tabular}
\end{adjustbox}
\caption{Ablation studies (using CUT as backbone) on the Label to Image task in defense of our choice in SRUNIT.}
\label{tab:2}
\end{table}


\vspace{-0.5 cm}

\section{Conclusion}
 In this paper, we tackle the semantic flipping problem in unpaired image translation which is critical for many of its applications.
 We argue that the inherently unmatched semantics distributions across different domains should be responded to by improving the semantic robustness of the generators.
 We do so by proposing a semantic robustness loss that enforces the semantics of the translated images to be invariant to the perceptual perturbations (specifically the multi-scale feature space perturbations) of the inputs.
 Quantitative and qualitative evaluations on multiple datasets suggest that our approach can effectively reduce semantics flipping that existing GAN-based methods suffer from.
 


{\small
\bibliographystyle{ieee_fullname}
\bibliography{egpaper_final}
}

\appendix

\section{Eqn. 3 vs. Eqn. 4 in the definition of $\mathcal{L}_{robust}$}
\label{app:eqn3v4}
In the main paper at Sec. 4.2, we briefly discuss the relation between the two versions (namely, Eqn. 3 \& 4) of $\mathcal{L}_{k}$ used in our proposed semantic robustness loss $\mathcal{L}_{robust}$.
Here we include more details.
To begin with, let us list them below as Eqn. \ref{eqn:a} and \ref{eqn:b}, respectively.

\begin{align} \label{eqn:a}
    \mathcal{L}_k = \mathbb{E}_x \Big[\frac{1}{||\tau_k||_2} \Big\Vert &F_k(\mathcal{G}_1^k(x)) \ - \\  &F_k(\mathcal{G}_1^k(\mathcal{G}_k^{K+1}(\mathcal{G}_1^k(x)+\tau_k))) \Big\Vert_2 \Big] \nonumber
\end{align}

\begin{align} \label{eqn:b}
    \mathcal{L}'_k = \mathbb{E}_x \Big[\frac{1}{||\tau_k||_2} \Big\Vert &F_k(\mathcal{G}_1^k(G(x))) \ - \\  &F_k(\mathcal{G}_1^k(\mathcal{G}_k^{K+1}(\mathcal{G}_1^k(x)+\tau_k))) \Big\Vert_2 \Big] \nonumber
\end{align}
The difference between the two is the extra $G(\cdot)$ inside the L2 norm of the Eqn. \ref{eqn:b}.
Remind that by the notations in the main paper we have $G(x) = \mathcal{G}_k^{K+1}(\mathcal{G}_1^k(x))$.
Optimizing Eqn. \ref{eqn:b} directly reflects our definition of semantic robustness in Sec. 3.4, i.e., the transformed image $G(x)$ should have their semantics computed by $F_k \circ \mathcal{G}_k$ invariant to small perturbations $\tau_k$ in the feature space (i.e. the output space of $G_k$) of the input $x$.
On the other hand, optimizing Eqn. \ref{eqn:a} indirectly minimizes Eqn. \ref{eqn:b}, since the contrastive loss (Eqn. 2 in Sec. 4,1 in the main paper) minimizes $|| F_k(\mathcal{G}_1^k(x)) - F_k(\mathcal{G}_1^k(G(x)))||_2$ due to all the outputs of $F_k$ constained to be on the same unit sphere (please refer to the CUT paper \cite{park2020contrastive} for details.)
In specific, denote $A_k(x) = F_k(\mathcal{G}_1^k(G(x)))$, $B_k(x) = F_k(\mathcal{G}_1^k(x))$ and $C_k(x) = F_k(\mathcal{G}_1^k(\mathcal{G}_k^{K+1}(\mathcal{G}_1^k(x)+\tau_k)))$, we have
\begin{align*}
    &\mathcal{L}'_k = \mathbb{E}_x \Big[\frac{1}{||\tau_k||_2} ||A_k(x) - C_k(x) ||_2 \big] \\
    &= \mathbb{E}_x \Big[\frac{1}{||\tau_k||_2} ||A_k(x) - B_k(x) + B_k(x) - C_k(x) ||_2 \big] \\
    &\le \mathbb{E}_x \Big[\frac{1}{||\tau_k||_2} \big(||A_k(x) - B_k(x)||_2 + ||B_k(x) - C_k(x)||_2\big) \Big] \\
\end{align*}
\begin{align*}
    &= \mathbb{E}_x ||A_k(x) - B_k(x)||_2  + \mathbb{E}_x \Big[\frac{1}{||\tau_k||_2}||B_k(x) - C_k(x)||_2 \Big]  \\
    &=  \mathbb{E}_x ||A_k(x) - B_k(x)||_2 + \mathcal{L}_k
\end{align*}
In short, we have
\[\mathcal{L}'_k \le \mathcal{L}_k + \mathbb{E}_x ||A_k(x) - B_k(x)||_2\]
Since $\frac{1}{K}\sum_{k=1}^K \mathbb{E}_x ||A_k(x) - B_k(x)||_2$ is minimized by the contrastive loss, our proposal to minimize $\mathcal{L}_{robust} = \frac{1}{K}\sum_{k=1}^K \mathcal{L}_k$ effectively minimizes an upper bound of $\frac{1}{K}\sum_{k=1}^K \mathcal{L}'_k$.  

We argue that optimizing such an upper bound (the adaptive version) is better than directly minimizing $\mathcal{L}'_k$ since otherwise the diversity of the translation can be harmed.
As a result, it might fail to produce complex visual patterns and create artifacts instead.
We numerically compare the performances in the ablation studies (Sec. 6 in the main paper).
We also show a visual comparison in Fig. \ref{fig:1}.

\begin{figure*}[t!]
	\centering
	\includegraphics[width=0.95\textwidth]{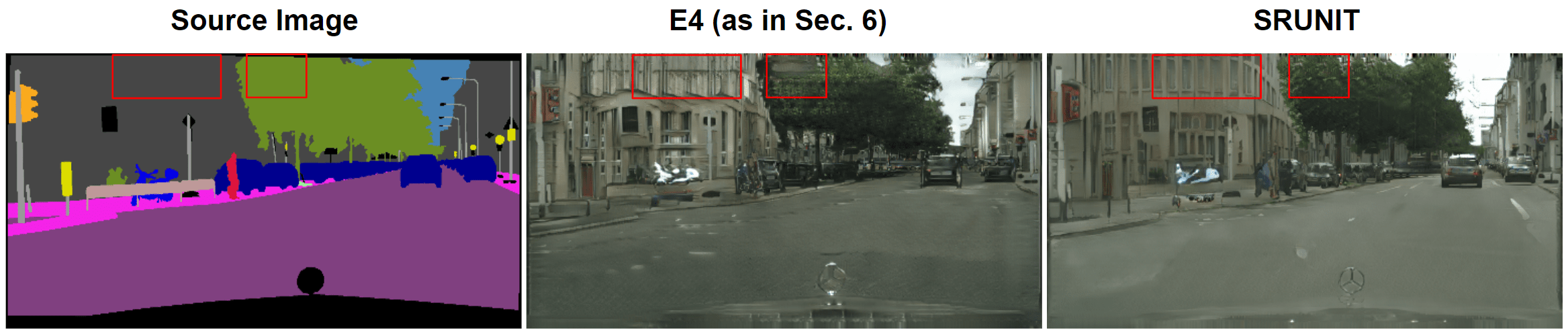}
	\caption{Visual comparisons of the Label to Image task for E4 (a model trained by using $\mathcal{L}'_k$ in optimizing $\mathcal{L}_{robust}$) vs. SRUNIT (by using $\mathcal{L}_k$ instead). The former does harm to the diversity of the translations. The red boxes highlight area where there are artifacts.}
    \label{fig:1}
\end{figure*}

\section{Evaluation on Label to Image and GTA to Cityscapes datasets}
When computing the three metrics on the Label to Image and the GTA to Cityscapes tasks, we use a light-weight DeepLab V3 \cite{chen2017rethinking} model pre-trained on the Cityscapes semantic segmentation task to evaluate the segmentation masks of the translated images.
In specific, we choose the \verb|mobilenetv3_small_cityscapes_trainfine| model, publicly available at
\href{https://github.com/tensorflow/models/blob/master/research/deeplab/g3doc/model_zoo.md}{TensorFlow Model Page}.

For both tasks and all 3 metrics, we perform the evaluation on the 500 (finely annotated) validation set of the Cityscapes dataset which consists of 19 classes.
We ignore all unlabelled pixels when computing the metrics.
We follow the common evaluation protocol as in \cite{park2020contrastive}.
For pxAcc (the average pixel accuracy), we compute the average pixel accuracy of the segmentation masks predicted by the DeepLab model on the translated images.
For clsAcc (the class accuracy), we compute the average pixel accuracy per semantic class and then compute the mean of those across all 19 classes.
For mIoU (mean IoU), we compute the average IoU per class and then the mean of them. 

\section{Dataset Construction with Unmatched Semantics Statistics}
\label{app:data}
Our paper focuses on unpaired image-to-image translation where data from two domains inherently have different semantics distributions.
One example where we quantitatively demonstrate this discrepancy is in Fig. 1 (from the main paper) for the GTA to Cityscapes task.
Since originally designed for paired image translations, the Label to Image and the Google Map to Aerial Photo datasets in our experiments are sub-sampled to ensure a reasonable amount of difference in their semantics statistics (briefly mentioned in Sec. 5.1.1, 5.1.3 \& 5.1.4).

In the Label to Image task from Cityscapes, as the original dataset is paired, we characterize each pair of images (the RGB semantic mask from the source domain and the street-view image from the target domain) by its histogram of the semantic classes (a vector $\in \mathbb{R}^{19}$, where the $i^{th}$ entry represents the ratio of the pixels belonging to the $i^{th}$ class in the image).
Then we use the K-means ($K = 2$) algorithm to cluster the 2975 images from the training set in Cityscapes according to these histograms (i.e., vectors).
We use all the source images from one cluster and all the target images from the other cluster as the sub-sampled unpaired data.
The resulting two clusters are of roughly the same size with different semantics statistics shown in Fig. \ref{fig:2} (top).

In the Google Map to Aerial Photo task (and vice versa), similarly, since the dataset is paired, we can characterize each out of the 1096 training pairs by the color histogram of its google map image.
In specific, we convert all google map image from RGB to gray-scale and apply bucketing to the pixel intensities (each bucket includes consecutively 5 out of the 256 total values) so that each histogram becomes a vector $\in \mathbb{R}^{51}$.
We apply K-means (again $K=2$) clustering on these histograms.
The resulting clusters are very different in size due to the long tail distribution in the Google Map dataset.
To deal with this, we first obtain the two histogram centroids from K-means.
Then, from highest to lowest, we rank all pairs of images by the ratio of the distance between its histogram to one centroid over that between its histogram to the other centroid.
We then assign the top half of the pairs to one cluster and the rest to the other.
Fig. \ref{fig:2} (bottom) shows the resulting distributions, where there is a large difference between the two.
We use all Google maps from cluster I together with 10\% of randomly sampled Google map images from cluster II (similarly all aerial photos from cluster II and 10\% from I) to ensure the reasonable amount of difference in semantics statistic between the two domains.

\begin{figure}[t!]
	\centering
	\includegraphics[width=0.45\textwidth]{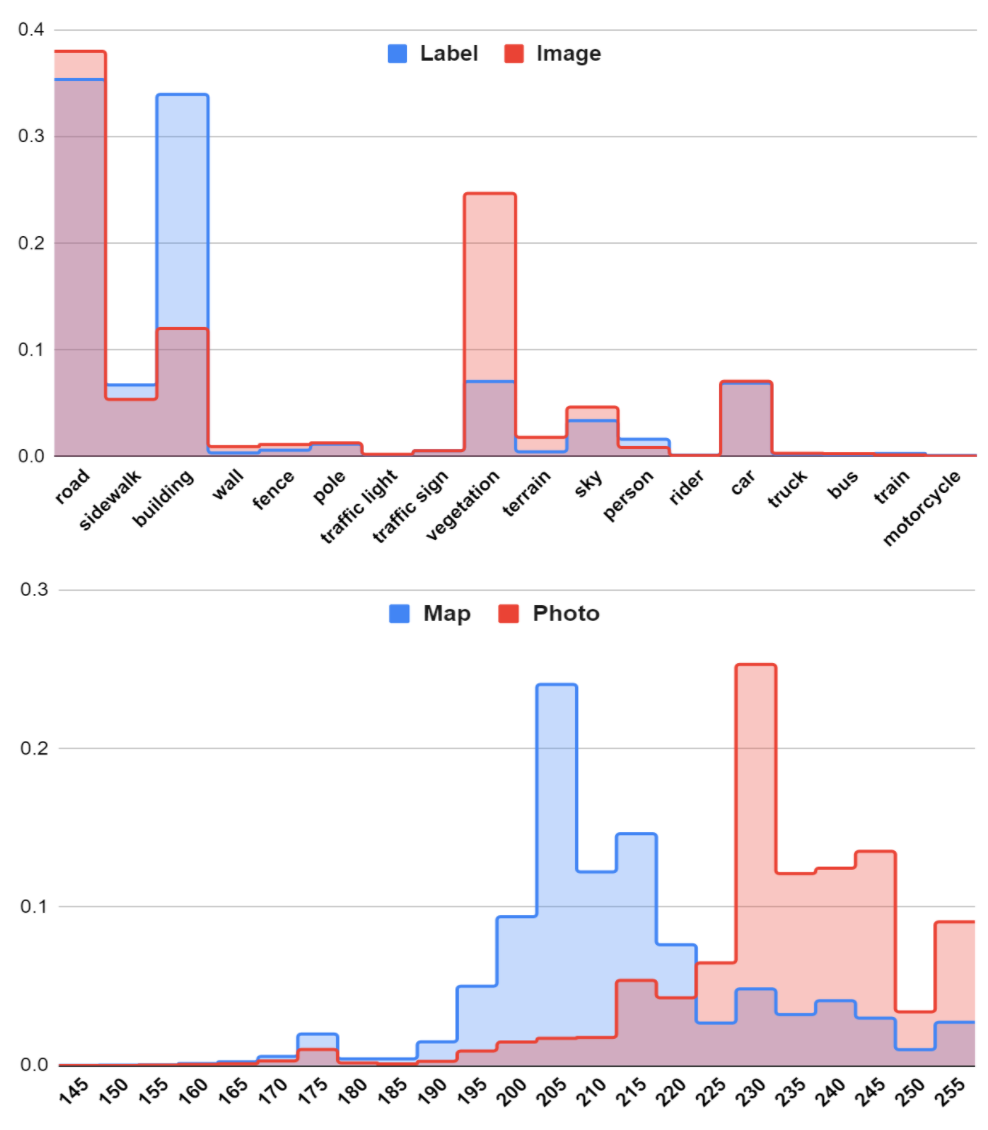}
	\caption{(\textbf{top}) Semantics distributions of the Label and Image data, each from one of the two clusters. (\textbf{bottom}) Semantics distributions of the Google Maps and Aerial Photos from the two clusters. As mentioned in Appendix \ref{app:data}, the latter has a very large semantics discrepancy and thus we mix 10\% of each cluster when constructing our sub-sampled dataset.}
    \label{fig:2}
\end{figure}

\section{Mismatched Semantics Statistics, Semantics Flipping \& Semantic Robustness}
In Sec. 3.2 of the main paper, we argue that it is inherently common in unpaired image translation to have mismatched semantics statistics across domains, and as a result, the semantics flipping usually occurs in the spurious solutions obtained by existing GAN-based methods.
We claim in Sec. 3.4 that our proposed semantics robustness can effectively mitigate the semantics flipping problems by, to some extent, offsetting the negative effects of the difference in semantics statistics.

\begin{figure}[t!]
	\centering
	\includegraphics[width=0.48\textwidth]{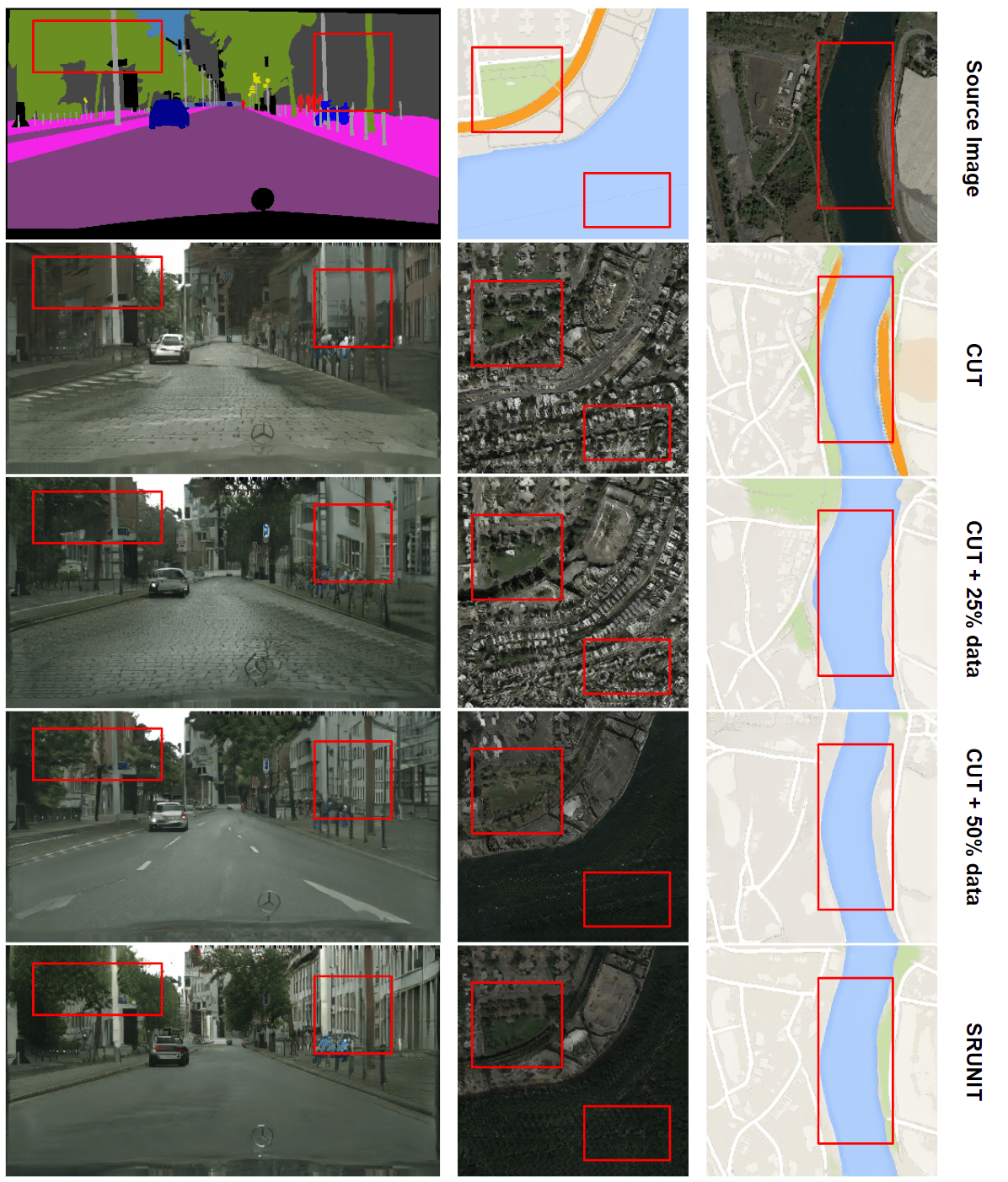}
	\caption{The three columns correspond to the experiments on Label to Image, Map to Photo and Photo to Map, respectively. CUT + X\% data indicates a CUT \cite{park2020contrastive} model trained with X\% more data sampled from the opposite cluster (to mitigate the mismatched semantics statistics problem). We shows that SRUNIT is rather effective in reducing semantics flipping caused by the different semantics statistics. It produces comparable or better translation results than the CUT baseline trained on data with much more ``matched'' semantics statistics. The red boxes highlight the areas where SRUNIT have improvements over CUT.}
    \label{fig:3}
\end{figure}

To give more insights, we construct training data with better-aligned semantics statistics between the source and target domains by adding more data to the training set from the opposite cluster.
For instance, in the Label to Image task where all source images are from cluster I and target ones from cluster II, we add X\% of source images randomly sampled from cluster II and X\% of target images from cluster I to form a new set of training data.
We compare several CUT \cite{park2020contrastive} models (the backbone of our proposed method SRUNIT) trained with data of different levels of discrepancy in semantics statistics.
The results (illustrated in Fig. \ref{fig:3}) indicate that (1) better-matched semantics statistics of the training data lead to less semantics flipping; (2) our method's improvement over its baseline in reducing semantics flipping is substantial.
As the results of SRUNIT are visually comparable to the baseline with 50\% more training data (which significantly reduces the difference in semantics statistics across the domains). 

\section{More Visual Results}
We show additional visual results for the 7 unpaired image-to-image translation tasks performed in the main paper.
They are Label to Image and GTA to Cityscapes (see Fig. \ref{fig:4}), Map to Photo and Photo to Map (see Fig. \ref{fig:5}), Horse to Zebra, Summer to Winter and Day to Night (see Fig. \ref{fig:6}), respectively.

\begin{figure*}[t!]
	\centering
	\includegraphics[width=0.95\textwidth]{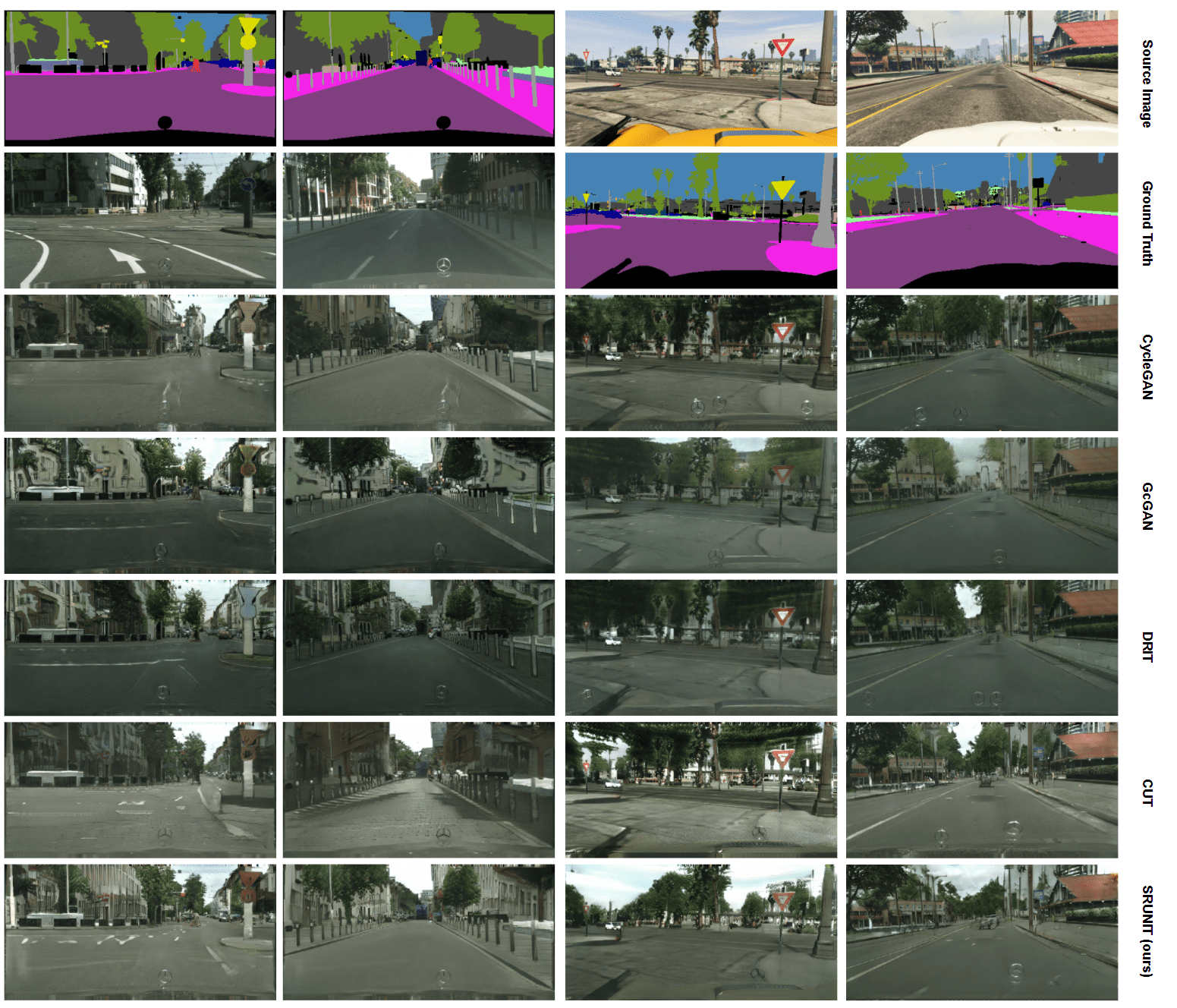}
	\caption{Additional visual results for the Label to Image and the GTA to Cityscapes tasks.}
    \label{fig:4}
\end{figure*}

\begin{figure*}[t!]
	\centering
	\includegraphics[width=0.95\textwidth]{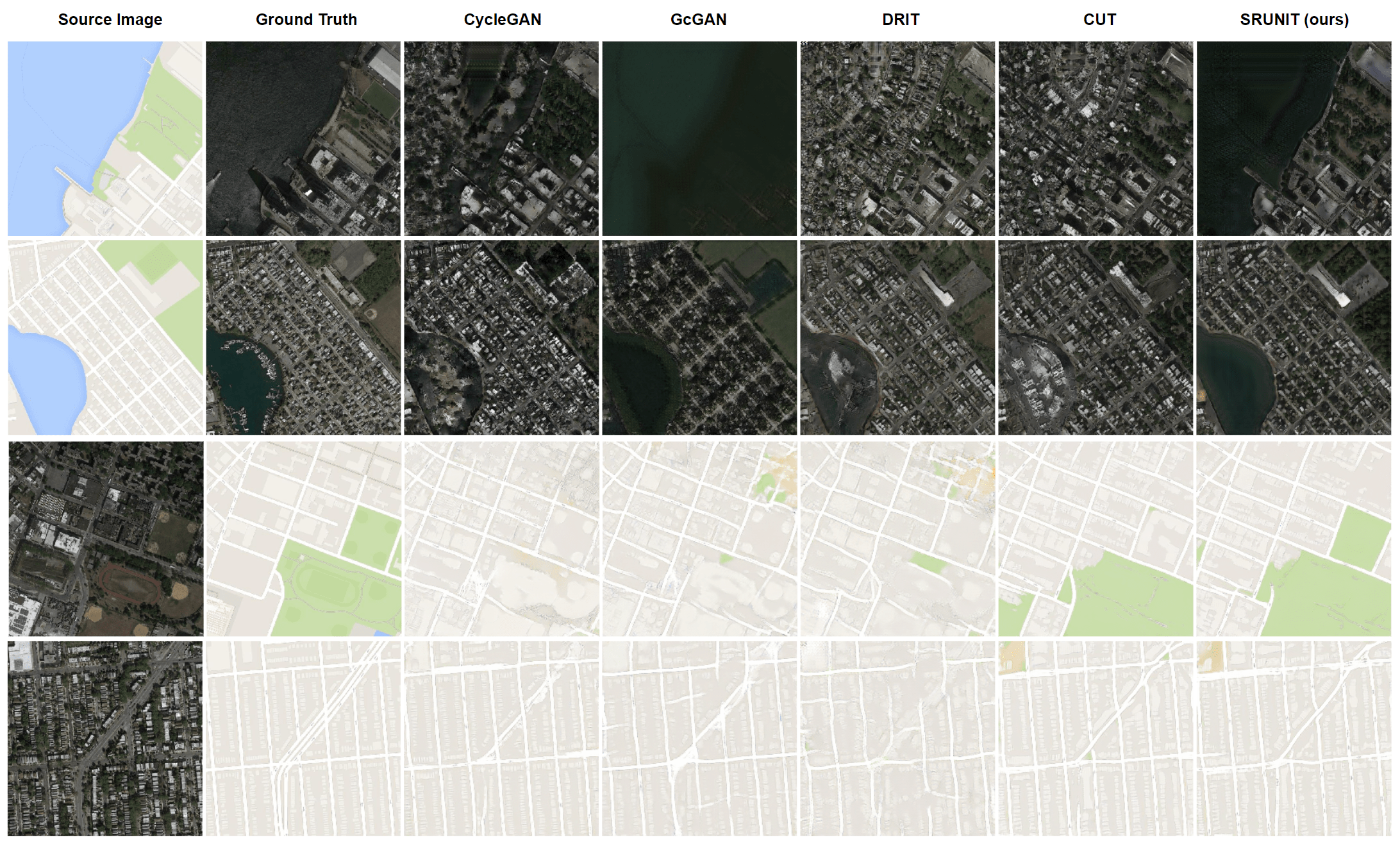}
	\caption{Additional visual results for the Map to Photo and the Photo to Map tasks.}
    \label{fig:5}
\end{figure*}

\begin{figure*}[t!]
	\centering
	\includegraphics[width=0.95\textwidth]{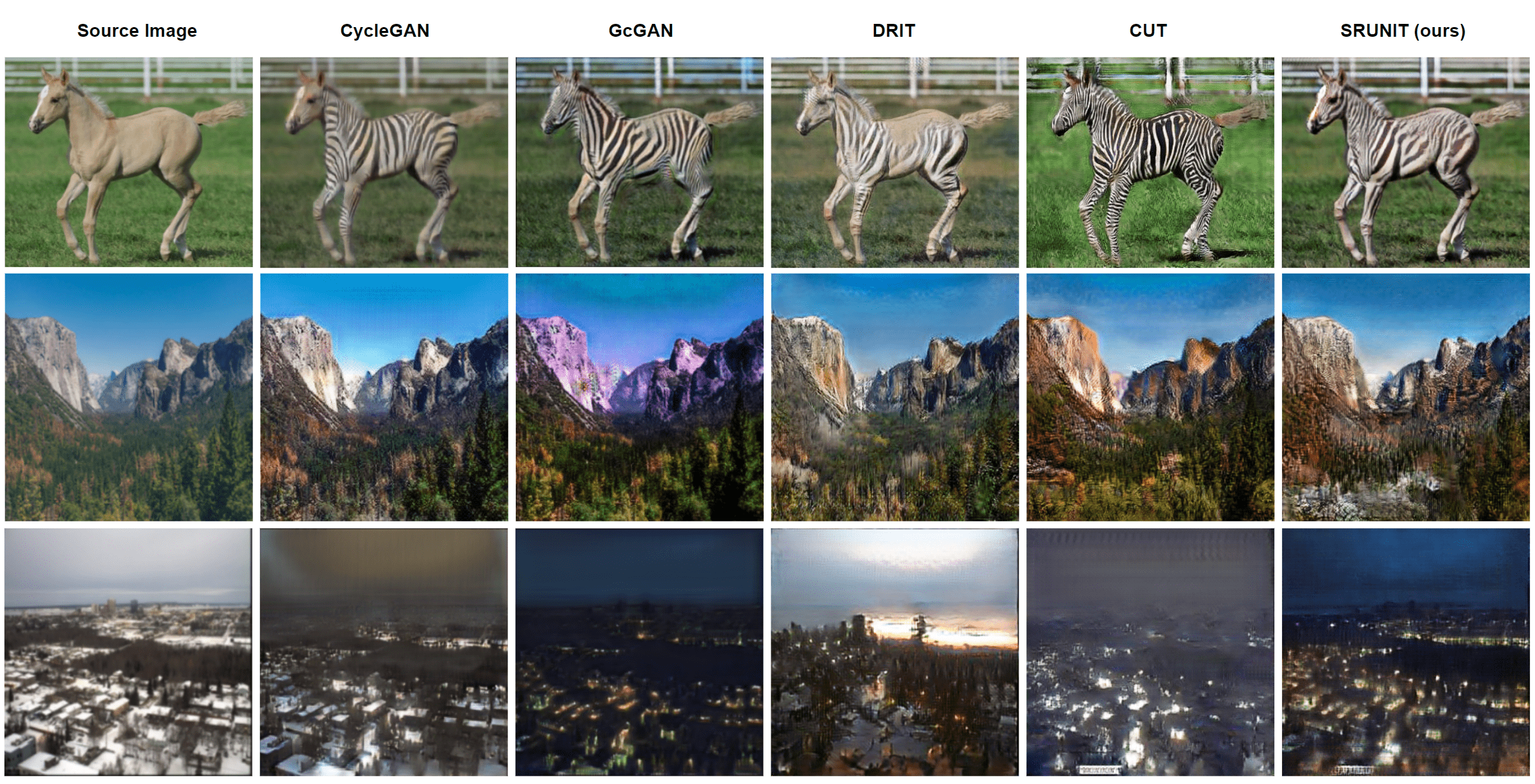}
	\caption{Additional visual results for the Horse to Zebra, the Summer to Winter and the Day to Night tasks.}
    \label{fig:6}
\end{figure*}

\section{Additional Implementation Details}
This section includes more implementation details about our method SRUNIT.
Please also refer to Sec. 5.3 of the main paper.
We follow CUT \cite{park2020contrastive} for the choice of network architecture and the training setup.
In specific, we use the least square loss \cite{mao2017least}, a ResNet-based generator \cite{johnson2016perceptual} with 9 residual blocks, and a patch-based discriminator \cite{isola2017image}.
We keep an image buffer of size 50 to update the discriminator for better training stability \cite{shrivastava2017learning}.
We use the default hyper-parameters for relevant loss terms in CUT, including selecting the same $K=5$ layers in the generator to compute the contrastive loss.
We add our proposed loss term $\beta * \mathcal{L}_{robust} = \frac{\beta}{K} \sum_{k=1}^K \mathcal{L}_k$ using the same $K$ layers (by default) with the coefficient $\beta = 10^{-4}$ (by default).
We fine-tune $K$ to be 4 or 5 by leaving one of the $\{mathcal{L}_k\}$ out each time.
We use Adam optimizers \cite{kingma2014adam} to train our model for 400 epochs with an initial learning rate of $0.0002$ and a linear decay for the last 50\% of epochs (the same with CUT).
The exception is that for the GTA to Cityscapes task, due to the large quantity of the training data, we only train for 20 epochs in total.
Moreover, we adopt the patch-based approach as in CUT to compute $\mathcal{L}_{robust}$; i.e., in each training iteration we randomly sample 256 patches from each of the $K$ layers to compute $\mathcal{L}_k$ (assuming batch size is 1).
This can significantly reduce the computation complexity in optimizing $\mathcal{L}_{robust}$.

\section{Details about the Ablation Studies}
For all models (E1 to E6) in the ablation studies section (Sec. 6 in the main paper), we use CUT as the backbone (the same as our proposed method SRUNIT).

\paragraph{E1} We aim to compare SRUNIT with the constraint proposed in DistanceGAN \cite{benaim2017one}.
E1 is trained by adding the self-distance loss given as:
\begin{equation*}
\begin{aligned}
\mathcal{L}_{\text {dist}}\left(G\right)=\mathbb{E}_{x} \big[& \frac{1}{\sigma_{X}}\left(\|L(x)-R(x)\|_{1}-\mu_{X}\right) \\
-& \frac{1}{\sigma_{Y}}\left(\left\|L\left(G(x)\right)-R\left(G(x)\right)\right\|_{1}-\mu_{Y}\right) \big]
\end{aligned}
\end{equation*}
where $L, R: \mathbb{R}^{H\times W\times 3} \rightarrow \mathbb{R}^{H\times \frac{W}{2}\times 3}$ are the operators that given an input image $x$ return the left or right part of it and $\sigma_{*}, \mu_{*}$ are the pre-computed image statistics from the two domain $X$ and $Y$ (see \cite{benaim2017one}).

\paragraph{E2} Similarly we compare SRUNIT with (a modified version of) the constraint proposed in HarmonicGAN \cite{zhang2019harmonic}.
We train E2 by adding the smoothness loss:
\begin{equation*}
\mathcal{L}_{smooth} = \mathbb{E}_{x_1, x_2} \big[ ||d(x_1, x_2) - d(G(x_1), G(x_2))||_1 \big]
\end{equation*}
where $x_1, x_2$ are image patches from the same input image, $d(\cdot, \cdot)$ is the distance function measured by using the histograms of two input patches, and $G(x_1)$ refers to image patch of the translation $G(x)$ corresponding to the patch $x_1$ from the input image $x$ (see details in \cite{zhang2019harmonic}).
In each iteration, we randomly sample 256 image patches to form 128 pairs from the input image to compute $\mathcal{L}_{smooth}$ (assuming the batch size is 1). 

\paragraph{E3} We aim to verify the necessity of using feature extractor $F_k$ in $\mathcal{L}_{robust}$.
We train E3 by removing $F_k$ in $\mathcal{L}_{robust}$, i.e., defining $\mathcal{L}_{robust} = \frac{1}{K}\sum_{k=1}^K \widetilde{\mathcal{L}}_k$, where 
\begin{align*}
     \widetilde{\mathcal{L}_k} = \mathbb{E}_x \Big[\frac{1}{||\tau_k||_2} \Big\Vert \mathcal{G}_1^k(x) - \mathcal{G}_1^k(\mathcal{G}_k^{K+1}(\mathcal{G}_1^k(x)+\tau_k)) \Big\Vert_2 \Big] 
\end{align*}

\paragraph{E4} We aim to empirically show the advantage of Eqn. \ref{eqn:a} over Eqn. \ref{eqn:b} (see Sec. 4.2 in the main paper and Appendix \ref{app:eqn3v4} for a discussion).
We train E4 by setting $\mathcal{L}_{robust} = \frac{1}{K}\sum_{k=1}^K \mathcal{L}'_k$ instead of $\frac{1}{K}\sum_{k=1}^K \mathcal{L}_k$.

\paragraph{E5} We aim to show that directly minimizing the distance between semantics extracted by $F_k$ of the input image and that of the corresponding translated image is not an effective way to reduce semantics flipping.
We train E5 by adding the semantics consistency term $\mathcal{L}^{sc} = \frac{1}{K}\sum_{k=1}^K \mathcal{L}_k^{sc}$, where
\begin{equation*}
\mathcal{L}^{sc}_k = \mathbb{E}_x \Vert F_k(\mathcal{G}_1^k(x)) - F_k(\mathcal{G}_1^k(G(x))) \Vert_2 
\end{equation*}
Since the semantics extractors $\{F_k\}$ are learned in an unsupervised manner, they are not accurate enough for direct enforcement of the preservation of semantics during the translations.

\paragraph{E6} Another direction of efforts to reduce semantics flipping is to pose constraints on the discriminator instead of on the generator.
We train E6 by applying the Lipschitz penalty \cite{gulrajani2017improved} to the discriminator in CUT.
Namely, we add the Lipschitz loss:
\begin{equation*}
\mathcal{L}_{lip} = \mathbb{E}_x [||\nabla_{x} D_Y(\overline{x})||_2 + ||\nabla_{\overline{G(x)}} D_Y(G(x))||_2]
\end{equation*}
where $\overline{x}$ is the mean value of $x$ and $\overline{G(x)}$ the mean value of $G(x)$.
We do so since CUT utilizes a patch-based discriminator and the gradient computation $\nabla$ is much cheaper than the Jacobian computation.

\paragraph{Remark:} In E3, E4, and E5, $x$ refers to the image patches instead of the entire images in the similar way as to how we adopt the patch-based approach in CUT to optimize our proposed semantic robustness loss $\mathcal{L}_{robust}$.

\end{document}